\documentclass[11pt, a4paper, logo, twocolumn, copyright, nonumbering]{googledeepmind}

\usepackage[authoryear, sort&compress, round]{natbib}
\usepackage[textwidth=18mm]{todonotes}
\usepackage{graphicx}
\usepackage{makecell}
\title{Gemma: Open Models Based on Gemini Research and Technology}

\renewcommand{\today}{2024-02-21}

\author{Gemma Team, Google DeepMind\authfootnotemark{1}}
\authfootnotetext{1}{See Contributions and Acknowledgments section for full author list. Please send correspondence to {\tt gemma-1-report@google.com}.}

\begin{abstract}
This work introduces Gemma, a family of lightweight, state-of-the art open models built from the research and technology used to create Gemini models. Gemma models demonstrate strong performance across academic benchmarks for language understanding, reasoning, and safety. We release two sizes of models (2 billion and 7 billion parameters), and provide both pretrained and fine-tuned checkpoints. Gemma outperforms similarly sized open models on 11 out of 18 text-based tasks, and we present comprehensive evaluations of safety and responsibility aspects of the models, alongside a detailed description of model development. We believe the responsible release of LLMs is critical for improving the safety of frontier models, and for enabling the next wave of LLM innovations.
\end{abstract}

\begin{document}

\maketitle

\section{Introduction}

We present Gemma, a family of open models based on Google's Gemini models \citep{geminiteam2023gemini}. 

We trained Gemma models on up to 6T tokens of text, using architectures, data, and training recipes inspired by the Gemini model family. Like Gemini, these models achieve strong generalist capabilities in text domains, alongside state-of-the-art understanding and reasoning skills at scale. With this work, we release both pre-trained and fine-tuned checkpoints, as well as an open-source codebase for inference and serving.

Gemma comes in two sizes: a 7 billion parameter model for efficient deployment and development on GPU and TPU, and a 2 billion parameter model for CPU and on-device applications. Each size is designed to address different computational constraints, applications, and developer requirements. At each scale, we release raw, pretrained checkpoints, as well as checkpoints fine-tuned for dialogue, instruction-following, helpfulness, and safety. We thoroughly evaluate the shortcomings of our models on a suite of quantitative and qualitative benchmarks. We believe the release of both pretrained and fine-tuned checkpoints will enable thorough research and investigation into the impact of current instruction-tuning regimes, as well as the development of increasingly safe and responsible model development methodologies.

Gemma advances state-of-the-art performance relative to comparable-scale (and some larger), open models \citep{mistral, llama2, llama1, falcon} across a wide range of domains including both automated benchmarks and human evaluation. Example domains include question answering \citep{boolq, natural-questions}, commonsense reasoning \citep{winogrande, bbhard}, mathematics and science \citep{gsm8k, mmlu}, and coding \citep{mbpp, humaneval}. See complete details in the \nameref{sec:evals} section.

\begin{figure*}[t]
\includegraphics[width=\textwidth]{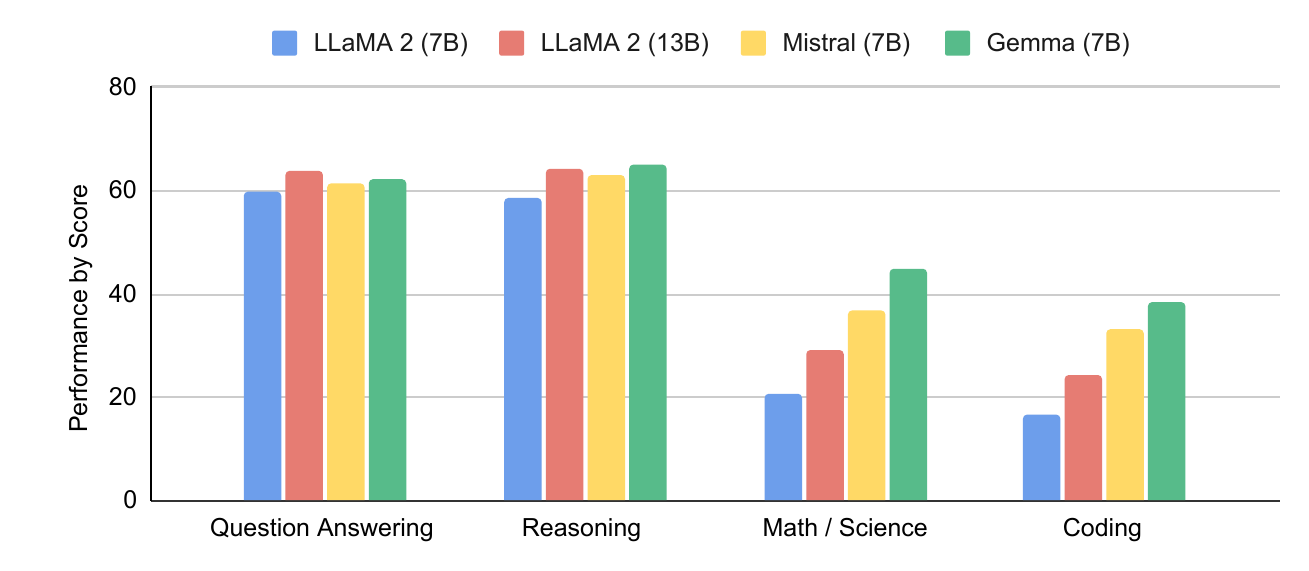}
\centering
\caption{Language understanding and generation performance of Gemma 7B across
different capabilities compared to similarly sized open models. We group together standard academic benchmark evaluations by capability and average the respective scores; see Table~\ref{tab:pretrain_auto_evals} for a detailed breakdown of performance.}
\label{fig:evals_chart}
\end{figure*}

Like Gemini, Gemma builds on recent work on sequence models \citep{DBLP:journals/corr/SutskeverVL14} and transformers \citep{DBLP:journals/corr/VaswaniSPUJGKP17}, deep learning methods based on neural networks \citep{deeplearningLeCun}, and techniques for large-scale training on distributed systems \citep{barham2022pathways, bradburyJAX, NIPS2012_6aca9700}. Gemma also builds on Google's long history of open models and ecosystems, including Word2Vec \citep{word2vec}, the Transformer \citep{DBLP:journals/corr/VaswaniSPUJGKP17}, BERT \citep{bert}, and T5 \citep{t5} and T5X \citep{t5x}.

We believe the responsible release of LLMs is critical for improving the safety of frontier models, for ensuring equitable access to this breakthrough technology, for enabling rigorous evaluation and analysis of current techniques, and for enabling the development of the next wave of innovations. While thorough testing of all Gemma models has been conducted, testing cannot cover all applications and scenarios in which Gemma may be used. With this in mind, all Gemma users should conduct rigorous safety testing specific to their use case before deployment or use. More details on our approach to safety can be found in section \nameref{safety}.

In this technical report, we provide a detailed overview of the model architecture, training infrastructure, and pretraining and fine-tuning recipes for Gemma, followed by thorough evaluations of all checkpoints across a wide-variety of quantitative and qualitative benchmarks, as well as both standard academic benchmarks and human-preference evaluations. We then discuss in detail our approach to safe and responsible deployment. Finally, we outline the broader implications of Gemma, its limitations and advantages.
\section{Model Architecture}
The Gemma model architecture is based on the transformer decoder \citep{DBLP:journals/corr/VaswaniSPUJGKP17}. The core parameters of the architecture are summarized in Table~\ref{tab:model_params}. Models are trained on a context length of 8192 tokens.
We also utilize several improvements proposed after the original transformer paper, and list them below:

\begin{table}
    \centering
    \begin{tabular}{l r r}
    \toprule
        Parameters & \textbf{2B} & \textbf{7B} \\
        \midrule
        \textit{d}\_{model} & 2048 & 3072 \\
        Layers & 18 & 28 \\
        Feedforward hidden dims & 32768 & 49152 \\
        Num heads & 8 & 16 \\
        Num KV heads & 1 & 16 \\
        Head size & 256 & 256 \\
        Vocab size & 256128 & 256128 \\
    \bottomrule
    \end{tabular}
    \caption{Key model parameters.}
    \label{tab:model_params}
\end{table}

\noindent\textbf{Multi-Query Attention} \citep{mqa}. 
Notably, the 7B model uses multi-head attention while the 2B checkpoints use multi-query attention (with $num\_kv\_heads = 1$), based on ablations that showed that multi-query attention works well at small scales \citep{mqa}. 

\noindent\textbf{RoPE Embeddings} \citep{rope}. Rather than using absolute positional embeddings, we use rotary positional embeddings in each layer; we also share embeddings across our inputs and outputs to reduce model size.

\noindent\textbf{GeGLU Activations} \citep{geglu}. The standard ReLU non-linearity is replaced by the approximated version of the GeGLU activation function.

\noindent\textbf{RMSNorm}. We normalize the input of each transformer sub-layer, the attention layer and the feedforward layer, with RMSNorm~\citep{rmsnorm} to stabilize the training.

\section{Training Infrastructure}

We train the Gemma models using TPUv5e; TPUv5e are deployed in pods of 256 chips, configured into a 2D torus of 16 x 16 chips. For the 7B model, we train our model across 16 pods, totaling to 4096 TPUv5e. 
We pretrain the 2B model across 2 pods, totaling 512 TPUv5e. 
Within a pod, we use 16-way model sharding and 16-way data replication for the 7B model. 
For the 2B, we simply use 256-way data replication. 
The optimizer state is further sharded using techniques similar to ZeRO-3. 
Beyond a pod, we perform data-replica reduce over the data-center network, using Pathways approach of \citep{barham2022pathways}.

\begin{table}
    \centering
    \begin{tabular}{l r r}
    \toprule
        Model & \makecell{Embedding\\Parameters} & \makecell{Non-embedding\\Parameters} \\
        \midrule
        \textbf{2B} & 524,550,144 &  1,981,884,416 \\
        \textbf{7B} & 786,825,216 & 7,751,248,896 \\
    \bottomrule
    \end{tabular}
    \caption{Parameter counts for the Gemma models. 
    We inherit from the large Gemini vocabulary (256k entries), that is designed to work on large quantities of languages, hence, the larger embedding parameter counts compared to models that are limited to one or a few languages.}
    \label{tab:model_param_counts}
\end{table}

We follow Gemini and we leverage the 'single controller' programming paradigm of Jax \citep{bradburyJAX} and Pathways \citep{barham2022pathways}. This simplifies the development process by enabling a single Python process to orchestrate the entire training run; we also leverage the GSPMD partitioner \citep{gspmd} for the training step computation and the MegaScale XLA compiler \citep{xla}.

\phantomsection
\subsection{Carbon Footprint}
We estimate the carbon emissions from pretraining the Gemma models to be $\sim131$ $tCO_2 eq$. This value is calculated based on the hourly energy usage reported directly from our TPU datacenters; we also scale this value to account for the additional energy expended to create and maintain the data center, giving us the total energy usage for our training experiments. We convert total energy usage to carbon emissions by joining our hourly energy usage against hourly per-cell carbon emission data reported by our data centers. 

In addition, Google data centers are carbon neutral, achieved through a combination of energy efficiency, renewable energy purchases, and carbon offsets. This carbon neutrality applies to our experiments and the machines running them. 

\phantomsection
\section{Pretraining}

\phantomsection
\subsection{Training Data}
Gemma 2B and 7B are trained on 3T and 6T tokens respectively of primarily-English data from web documents, mathematics, and code. Unlike Gemini, these models are not multimodal, nor are they trained for state-of-the-art performance on multilingual tasks.

We use a subset of the SentencePiece tokenizer \citep{kudo-richardson-2018-sentencepiece} of Gemini for compatibility.  It splits digits, does not remove extra whitespace, and relies on byte-level encodings for unknown tokens, following the techniques used for both \citep{chowdhery2022palm} and \citep{geminiteam2023gemini}. The vocabulary size is 256k tokens.

\phantomsection
\subsection{Filtering}

We filter the pre-training dataset to reduce the risk of unwanted or unsafe utterances, and filter out certain personal information or other sensitive data. This includes both heuristics and model-based classifiers to remove harmful or low-quality content. Further, we filter all evaluation sets from our pre-training data mixture, run targeted contamination analyses to check against evaluation set leakage, and reduce the risk of recitation by minimizing proliferation of sensitive outputs. 

The final data mixture was determined through a series of ablations on both the 2B and 7B models. Similar to the approach advocated in \citep{geminiteam2023gemini}, we stage training to alter the corpus mixture throughout training to increase the weight of relevant, high-quality data towards the end of training.

\section{Instruction Tuning}

We finetune Gemma 2B and 7B with supervised fine-tuning (SFT) on a mix of text-only, English-only synthetic and human-generated prompt-response pairs and reinforcement learning from human feedback (RLHF) with the reward model trained on labelled English-only preference data and the policy based on a set of high-quality prompts. We find that both stages are important for improved performance on downstream automatic evaluations and human preference evaluations of model outputs.

\phantomsection
\subsection{Supervised Fine-Tuning}

We selected our data mixtures for supervised fine-tuning based on LM-based side-by-side evaluations \citep{zheng2023judging}. 
Given a set of held-out prompts, we generate responses from a test model, generate responses on the same prompts from a baseline model, shuffle these randomly, and ask a larger, high capability model to express a preference between two responses. Different prompt sets are constructed to highlight specific capabilities, such as instruction following, factuality, creativity, and safety. Our LM-based judges employ a number of known strategies, such as chain-of-thought prompting \citep{chain-of-thought}, rubrics and constitutions \citep{bai2022constitutional}, to be aligned with human preferences.

\phantomsection
\subsection{Filtering}
When using synthetic data, we run several stages of filtering over it, removing examples that show certain personal information, unsafe or toxic model outputs, mistaken self-identification data, or duplicated examples. Following Gemini, we find that including subsets of data that encourage better in-context attribution, hedging, and refusals to minimize hallucinations improves performance on factuality metrics, without degrading model performance on other metrics. 

The final data mixtures and supervised fine-tuning recipe, which includes tuned hyperparameters, were chosen on the basis of improving helpfulness while minimizing model harms related to safety and hallucinations. 

\phantomsection
\subsection{Formatting}
Instruction tuned models are trained with a specific formatter that annotates all instruction tuning examples with extra information, both at training and inference time. It has two purposes: 1) indicating roles in a conversation, such as the User role, and 2) delineating turns in a conversation, especially in a multi-turn conversation. Special control tokens are reserved in the tokenizer for this purpose. While it is possible to get coherent generations without the formatter, it will be out-of-distribution for the model, and will very likely produce worse generations.

The relevant formatting control tokens are presented in Table \ref{tab:formatting_tokens}, with a dialogue example presented in Table \ref{tab:sample_dialogue}.

\begin{table}[ht!]
    \setlength{\tabcolsep}{6pt}
    \centering
    \footnotesize
    \begin{tabular}{l c c}
    \toprule
    \textbf{Context} & \textbf{Relevant Token} \\
        \midrule
        \scriptsize{User turn} & \texttt{\color{NavyBlue}user} \\
        \midrule
        \scriptsize{Model turn} & \texttt{\color{NavyBlue}model} \\
        \midrule
        \scriptsize{Start of conversation turn} & \texttt{\color{NavyBlue}<start\_of\_turn>} \\
        \midrule
        \scriptsize{End of conversation turn} & \texttt{\color{NavyBlue}<end\_of\_turn>} \\
    \bottomrule
    \end{tabular}
    \caption{Relevant formatting control tokens used for both SFT and RLHF of Gemma models.}
    \label{tab:formatting_tokens}
\end{table}

\begin{table}[ht!]
    \setlength{\tabcolsep}{6pt}
    \centering
    \footnotesize   
    \begin{tabular}{r l}
    \toprule
    \vspace{0.2cm}
    \textbf{User:} & {\color{NavyBlue}\texttt{<start\_of\_turn>user}} \vspace{-0.2cm} \\
    & \texttt{Knock knock.}{\color{NavyBlue}\texttt{<end\_of\_turn>}} \\
    & {\color{NavyBlue}\texttt{<start\_of\_turn>model}} \vspace{0.1cm} \\
    
    \textbf{Model:} & \texttt{Who's there?}{\color{NavyBlue}\texttt{<end\_of\_turn>}} \vspace{0.1cm} \\

    \textbf{User:} & {\color{NavyBlue}\texttt{<start\_of\_turn>user}} \\
    & \texttt{Gemma.}{\color{NavyBlue}\texttt{<end\_of\_turn>}} \\
    & {\color{NavyBlue}\texttt{<start\_of\_turn>model}} \vspace{0.1cm} \\

    \textbf{Model:} & \texttt{Gemma who?}{\color{NavyBlue}\texttt{<end\_of\_turn>}} \vspace{0.1cm} \\

    \bottomrule
    \end{tabular}
    \caption{Example dialogue with user and model control tokens.}
    \label{tab:sample_dialogue}
    \vspace{-0.5cm}
\end{table}

\subsection{Reinforcement Learning from Human Feedback}

We further finetuned the supervised fine-tuned model using RLHF \citep{christiano2017deep,ouyang2022training}. We collected pairs of preferences from human raters and trained a reward function under the Bradley-Terry model \citep{bradley1952rank}, similarly to Gemini. The policy was trained to optimize this reward function using a novel reinforcement learning algorithm. Similar to the SFT phase, and in order to tune hyperparameters and additionally mitigate reward hacking \citep{amodei2016concrete,skalse2022defining} we relied on a high capacity model as an automatic rater and computed side-by-side comparisons against baseline models.

\phantomsection
\section{Evaluation}
\label{sec:evals}

We evaluate Gemma across a broad range of domains, using both automated benchmarks and human evaluation.

\phantomsection
\subsection{Human Preference Evaluations}
\label{sec:humanevals}

In addition to running standard academic benchmarks on the finetuned models, we sent final release candidates to human evaluation studies to be compared against the Mistral v0.2 7B Instruct model \citep{mistral}.

On a held-out collection of around 1000 prompts oriented toward asking models to follow instructions across creative writing tasks, coding, and following instructions, Gemma 7B IT has a 61.2\% positive win rate and Gemma 2B IT has a 45\% win rate over Mistral v0.2 7B Instruct. On a held-out collection of around 400 prompts oriented towards testing basic safety protocols, Gemma 7B IT has a 63.5\% win rate, while Gemma 2B IT has a 60.1\% win rate. We report the corresponding numbers in Table~\ref{tab:it}.

\begin{table}[t!]
    \setlength{\tabcolsep}{6pt}
    \centering
    \footnotesize
    \begin{tabular}{l r r}
    \toprule
    Model & Safety & Instr. Following \\
    \midrule
   \textbf{Gemma 1.1 IT 7B} & \textbf{63.5\%} &  \textbf{61.2\%} \\
\tiny{\textit{95\% Conf. Interval}} & \tiny{[60.7\%, 66.1\%]} & \tiny{[59.3\%, 63\%]} \vspace{-0.05cm} \\
\tiny{\textit{Win / Tie / Loss}} & \tiny{51.5\% / 23.9\% / 24.6\%} & \tiny{52.2\% / 18.1\% / 29.8\%} \vspace{0.2cm} \\
   \textbf{Gemma 1.1 IT 2B} & \textbf{60.1\%} &  45\% \\
\tiny{\textit{95\% Conf. Interval}} & \tiny{[57.3\%, 62.8\%]} & \tiny{[43.1\%, 46.9\%]}  \vspace{-0.05cm} \\
\tiny{\textit{Win / Tie / Loss}} & \tiny{48.5\% / 23.2\% / 28.3\%} & \tiny{37.1\% / 15.8\% / 47.1\%} \\
    \bottomrule
    \end{tabular}
    \caption{Win rate of Gemma 1.1 IT models versus Mistral 7B v0.2 Instruct with 95\% confidence intervals. We report breakdowns of wins, ties, and losses, and we break ties evenly when reporting the final win rate. Gemma 1.0 results can be found in the appendix.}
    \label{tab:it}
\end{table}

\phantomsection
\subsection{Automated Benchmarks}

\begin{table*}[t!]
    \centering
    \begin{tabular}{l c c c c c c }
    \toprule
         & & \multicolumn{2}{c}{LLaMA-2} & Mistral & \multicolumn{2}{c}{Gemma}\\
         \cmidrule(l{3pt}r{3pt}){3-4}\cmidrule(l{3pt}r{3pt}){5-5}\cmidrule(l{3pt}r{3pt}){6-7}
Benchmark & ~~~~~~~~~~~~metric~~~~~~~~~~~~  & 7B & 13B & 7B & 2B & 7B \\
      \midrule
MMLU & 5-shot, top-1 & 45.3 & 54.8 & 62.5 & 42.3 & \textbf{64.3} \\
        \midrule
HellaSwag & 0-shot   & 77.2 & 80.7 & 81.0 & 71.4 & \textbf{81.2}\\
PIQA & 0-shot        & 78.8 & 80.5 & \textbf{82.2} & 77.3 & 81.2\\
SIQA & 0-shot       & 48.3 & 50.3 &  \phantom{$^*$}47.0$^*$ & 49.7 & \textbf{51.8}  \\
Boolq & 0-shot      & 77.4 & 81.7 &  \phantom{$^*$}\textbf{83.2}$^*$ & 69.4 & \textbf{83.2}  \\
Winogrande & partial scoring & 69.2 & 72.8 & \textbf{74.2} & 65.4 & 72.3  \\
CQA & 7-shot  & 57.8 & 67.3 & \phantom{$^*$}66.3$^*$ & 65.3 & \textbf{71.3}  \\
OBQA & & \textbf{58.6} & 57.0 & 52.2 &  47.8 & 52.8  \\
ARC-e & & 75.2 & 77.3 & 80.5 & 73.2 & \textbf{81.5} \\
ARC-c &  & 45.9 & 49.4 & \textbf{54.9} & 42.1 & 53.2\\
\midrule
TriviaQA & 5-shot & 72.1 & \textbf{79.6} & 62.5 & 53.2 & 63.4 \\
NQ & 5-shot & 25.7 & \textbf{31.2} & 23.2 &  12.5 & 23.0 \\
\midrule
HumanEval & pass@1 & 12.8 & 18.3 & 26.2 & 22.0 & \textbf{32.3} \\
MBPP$^{\dagger}$ & 3-shot & 20.8 & 30.6 & \phantom{$^*$}40.2$^*$ & 29.2 & \textbf{44.4}  \\
GSM8K & maj@1 & 14.6 & 28.7 &  \phantom{$^*$}35.4$^*$ & 17.7 & \textbf{46.4}  \\
MATH & 4-shot  & 2.5 & 3.9 & 12.7 & 11.8 & \textbf{24.3} \\
\midrule
AGIEval & & 29.3 & 39.1 &  \phantom{$^*$}41.2$^*$ & 24.2 & \textbf{41.7} \\
BBH & & 32.6 & 39.4 & \phantom{$^*$}\textbf{56.1}$^*$ & 35.2 & 55.1  \\
\midrule
Average & & 46.9 & 52.4 & 54.5 & 45.0 & \textbf{56.9}  \\
\bottomrule
\end{tabular}
    \caption{Academic benchmark results, compared to similarly sized, openly-available models trained on general English text data.
   $^{\dagger}$ Mistral reports 50.2 on a different split for MBPP and on their split our 7B model achieves 54.5.
   $^*$ evaluations run by us. Note that due to restrictive licensing, we were unable to run evals on LLaMA-2; all values above were previously reported in \cite{llama2}.
    }
    \label{tab:pretrain_auto_evals}
\end{table*}

We measure Gemma models' performance on domains including physical reasoning \citep{piqa}, social reasoning \citep{siqa}, question answering \citep{boolq,natural-questions}, coding \citep{mbpp,humaneval}, mathematics \citep{gsm8k}, commonsense reasoning \citep{winogrande}, language modeling \citep{lambada}, reading comprehension \citep{triviaqa}, and more. 

\begin{table}[ht]
    \centering
    \begin{tabular}{l c c}
    \toprule
         &  Mistral & Gemma\\
Benchmark   &  7B & 7B \\
      \midrule
ARC-c     &   60.0 & \textbf{61.9}\\
HellaSwag &   \textbf{83.3} & 82.2  \\
MMLU      &   64.2 & \textbf{64.6}\\
TruthfulQA &   42.2 & \textbf{44.8} \\
Winogrande &   78.4 &  \textbf{79.0} \\
GSM8K      &   37.8 & \textbf{50.9}   \\
\midrule
Average &  61.0 & \textbf{63.8} \\
\bottomrule
\end{tabular}
    \caption{HuggingFace H6 benchmark. 
    The performance of small models are sensitive to small modifications in prompts and we further validate the quality of our models on an independent implementation of multiple known benchmarks. All evaluations were run by HuggingFace.}
    \label{tab:pretrain_auto_evals_gemini_caption}
\end{table}

For most automated benchmarks we use the same evaluation methodology as in Gemini. Specifically for those where we report performance compared with Mistral, we replicated methodology from the Mistral technical report as closely as possible. These specific benchmarks are: ARC \citep{arc}, CommonsenseQA \citep{commonsenseqa},  Big Bench Hard \citep{bbhard}, and AGI Eval (English-only) \citep{zhong2023agieval}. Due to restrictive licensing, we were unable to run any evaluations on LLaMA-2 and cite only those metrics previously reported \citep{llama2}.

We compare Gemma 2B and 7B models to several external open-source (OSS) LLMs across a series of academic benchmarks, reported in Table \ref{tab:pretrain_auto_evals} and Table~\ref{tab:pretrain_auto_evals_gemini_caption}.

On MMLU \citep{mmlu}, Gemma 7B outperforms all OSS alternatives at the same or smaller scale; it also outperforms several larger models, including LLaMA2 13B. However, human expert performance is gauged at 89.8\% by the benchmark authors; as Gemini Ultra is the first model to exceed this threshold, there is significant room for continued improvements to achieve Gemini and human-level performance.

Gemma models demonstrate particularly strong performance on mathematics and coding benchmarks. On mathematics tasks, which are often used to benchmark the general analytical capabilities of models, Gemma models outperform other models by at least 10 points on GSM8K \citep{gsm8k} and the more difficult MATH \citep{hendrycksmath2021} benchmark. Similarly, they outperform alternate open models by at least 6 points on HumanEval \citep{humaneval}. They even surpass the performance of the code-fine-tuned CodeLLaMA-7B models on MBPP (CodeLLaMA achieves a score of 41.4\% where Gemma 7B achieves 44.4\%). 

\begin{table*}
    \centering
    \begin{tabular}{l c c c  c c  c }
    \toprule
         & & Mistral v0.2 & \multicolumn{2}{c}{Gemma 1.1 IT}\\
        \cmidrule(l{3pt}r{3pt}){3-3}\cmidrule(l{3pt}r{3pt}){4-5}
Benchmark & metric  & 7B* & 2B & 7B \\
      \midrule
RealToxicity & avg & 8.44 & \textbf{7.03} & 8.04 \\
BOLD &  & 46.0 & \textbf{47.76} & 45.2  \\
CrowS-Pairs & top-1 & 32.76 & 45.89 & \textbf{49.67}  \\
BBQ Ambig & 1-shot, top-1 & \textbf{97.53} & 58.97 & 86.06 \\
BBQ Disambig & top-1 & 84.45 & 53.9 & \textbf{85.08} \\
Winogender & top-1 & \textbf{64.3} & 50.14 & 57.64 \\
TruthfulQA & & \textbf{48.54} & 44.24 & 45.34\\
Winobias 1\_2 &  & \textbf{65.72} & 55.93 & 59.22 \\
Winobias 2\_2 &  & 84.53 & \textbf{89.46} & 89.2 \\
Toxigen &  & 61.77 & \textbf{29.64} & 38.75 \\
\bottomrule
\end{tabular}
    \caption{Safety academic benchmark results of Gemma 1.1 IT models, compared to similarly sized, openly-available models. Evaluations run by us. Note that due to restrictive licensing, we were unable to run evals on LLaMA-2; we do not report previously-published numbers for LLaMA-2  on TruthfulQA, as we use different, non-comparable evaluation set-ups: we use MC2, where LLaMA-2 uses GPT-Judge. Results for Gemma 1.0 IT models can be found in appendix.
    }
    \label{tab:safety_auto_evals}
\end{table*}

\subsection{Memorization Evaluations}\label{sec:memorization}

Recent work has shown that aligned models may be vulnerable to new adversarial attacks that can bypass alignment \citep{nasr2023scalable}. These attacks can cause models to diverge, and sometimes regurgitate memorized training data in the process. We focus on discoverable memorization, which serves as a reasonable upper-bound on the memorization of a model~\citep{nasr2023scalable} and has been the common definition used in several studies~\citep{carlini2022quantifying,anil2023palm,kudugunta2023madlad}.

\begin{figure}[htb!]
    \centering
    \includegraphics[width=\linewidth]{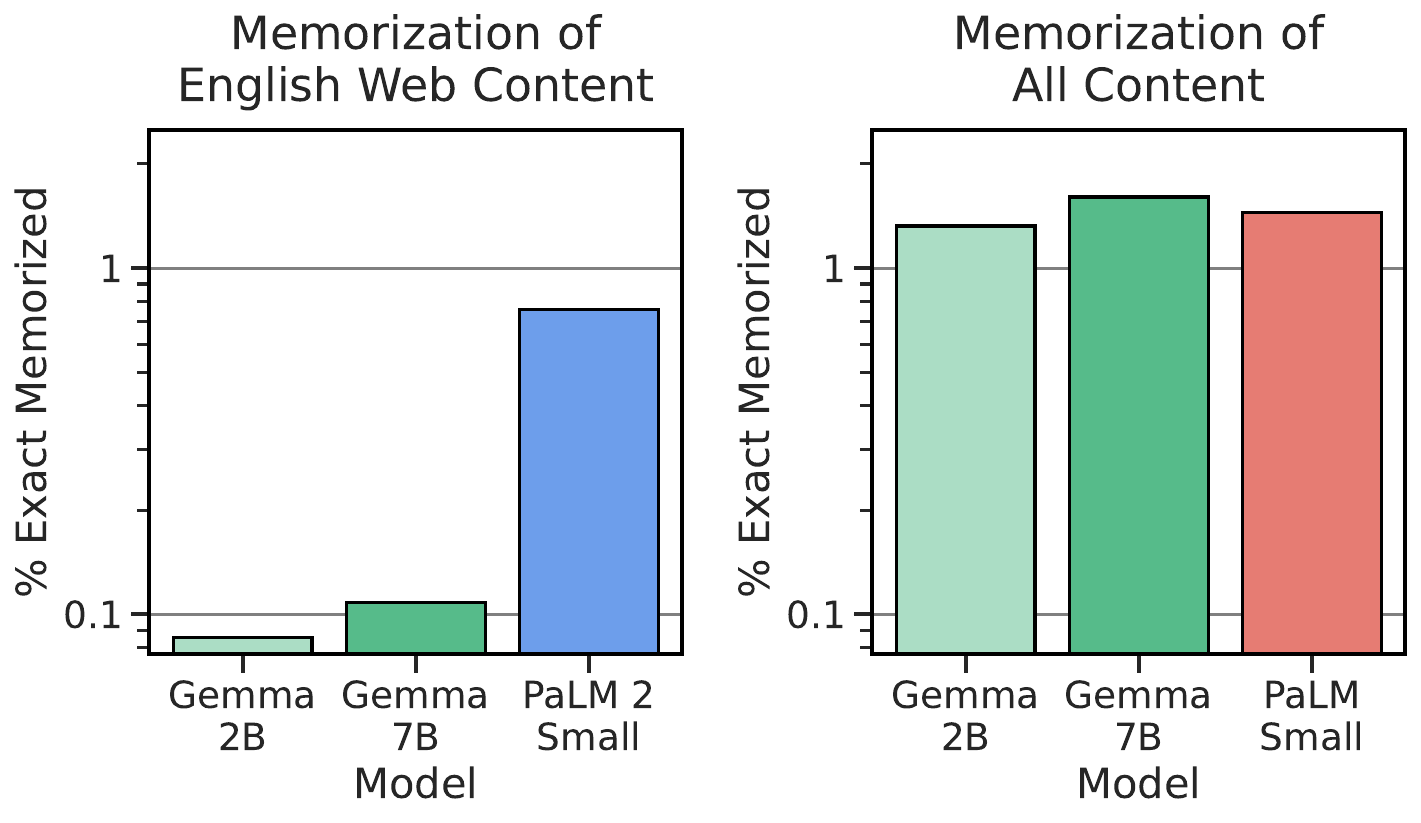}
    \caption{Comparing average memorization rates across model families. We compare the Gemma pretrained models to PaLM and PaLM 2 models of comparable size and find similarly low rates of memorization.}
    \label{fig:mem-size}
\end{figure}

We test for memorization\footnote{Our use of “memorization” relies on the definition of that term found at www.genlaw.org/glossary.html.} of the Gemma pretrained models with the same methodology performed in~\citet{anil2023palm}. We sample 10,000 documents from each corpus and use the first 50 tokens as a prompt for the model. We focus mainly on exact memorization, where we classify texts as memorized if the subsequent 50 tokens generated by the model exactly match the ground truth continuation in the text. However, to better capture potential paraphrased memorizations, we include approximate memorization~\citep{ippolito2022preventing} using an 10\% edit distance threshold. In Figure~\ref{fig:mem-size}, we compare the results of our evaluation with the closest sized PaLM~\citep{chowdhery2022palm} and PaLM 2 models~\citep{anil2023palm}.

\paragraph{Verbatim Memorization} PaLM 2 compared with PaLM by evaluating on a shared subset of their training corpora. However, there is even less overlap between the Gemma pretraining data with the PaLM models, and so using this same methodology, we observe much lower memorization rates (Figure~\ref{fig:mem-size} left). Instead, we find that estimating the ``total memorization'' across the entire pretraining dataset gives a more reliable estimate (Figure~\ref{fig:mem-size} right) where we now find the Gemma memorizes training data at a comparable rate to PaLM. 

\begin{figure}[htb!]
    \centering
    \includegraphics[width=\linewidth]{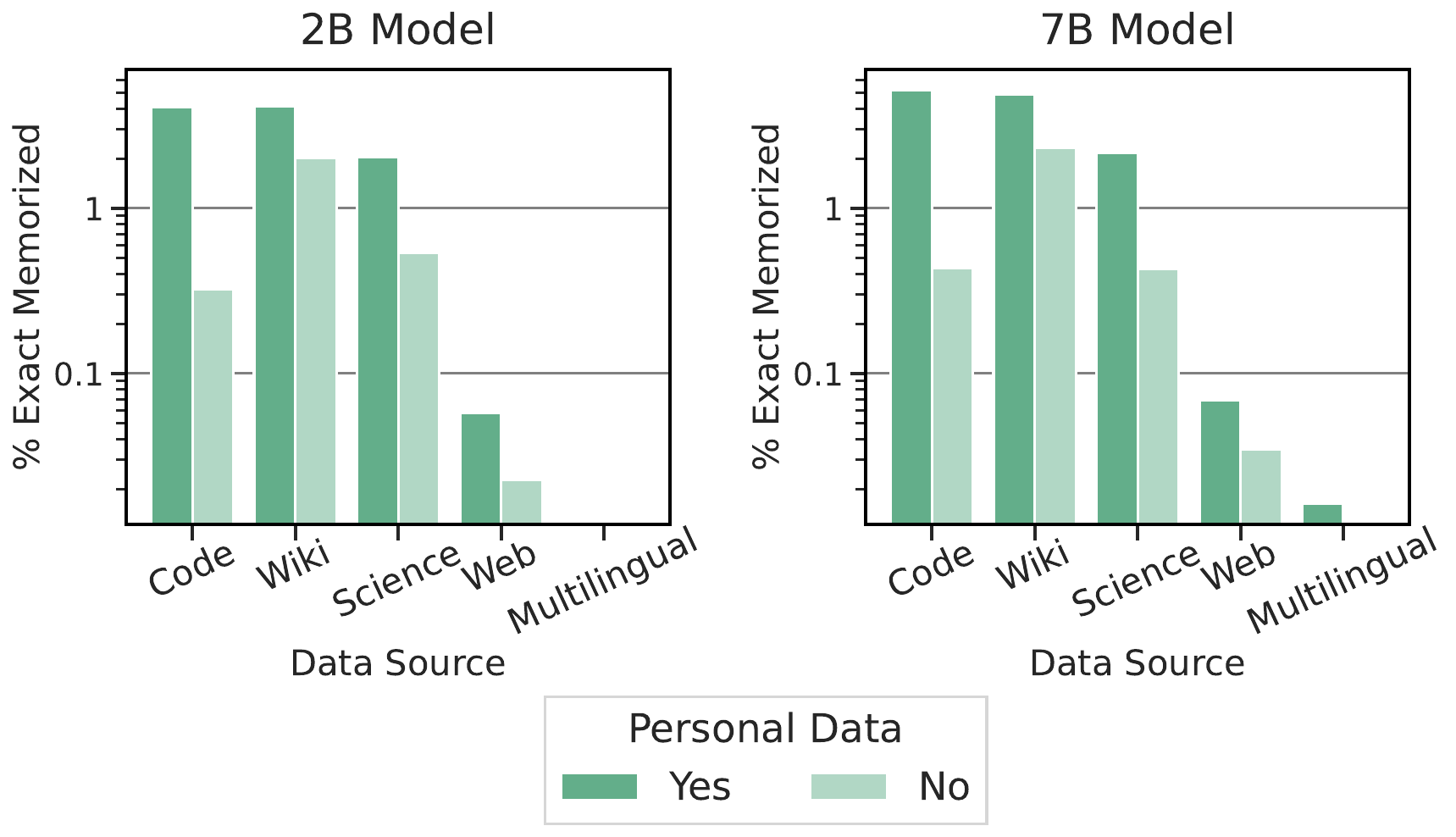}
    \caption{Measuring personal and sensitive data memorization rates. \textbf{No sensitive data was memorized, hence it is omitted from the figure}.}
    \label{fig:mem-sensitive}
\end{figure}

\paragraph{Personal Data}

Perhaps of higher importance is the possibility that personal data might be memorized. As part of making Gemma pre-trained models safe and reliable, we used automated techniques to filter out certain personal information and other sensitive data from training sets.  

To identify possible occurrences of personal data, we use Google Cloud Sensitive Data Protection\footnote{Available at: \url{https://cloud.google.com/sensitive-data-protection}}. This tool outputs three severity levels based on many categories of personal data (e.g., names, emails, etc.). We classify the highest severity as “sensitive” and the remaining two as simply “personal”. Then, we measure how many memorized outputs contain any sensitive or personal data. As shown in Figure~\ref{fig:mem-sensitive}, \emph{we observe no cases of memorized sensitive data.} We do find that the model memorizes some data we have classified as potentially “personal” according to the above, though often at a much lower rate.  Further, it is important to note that these tools are known to have many false positives (because they only match patterns and do not consider the context), meaning that our results are likely overestimates of the amount of personal data identified.

\begin{figure}[htb!]
    \centering
    \includegraphics[width=\linewidth]{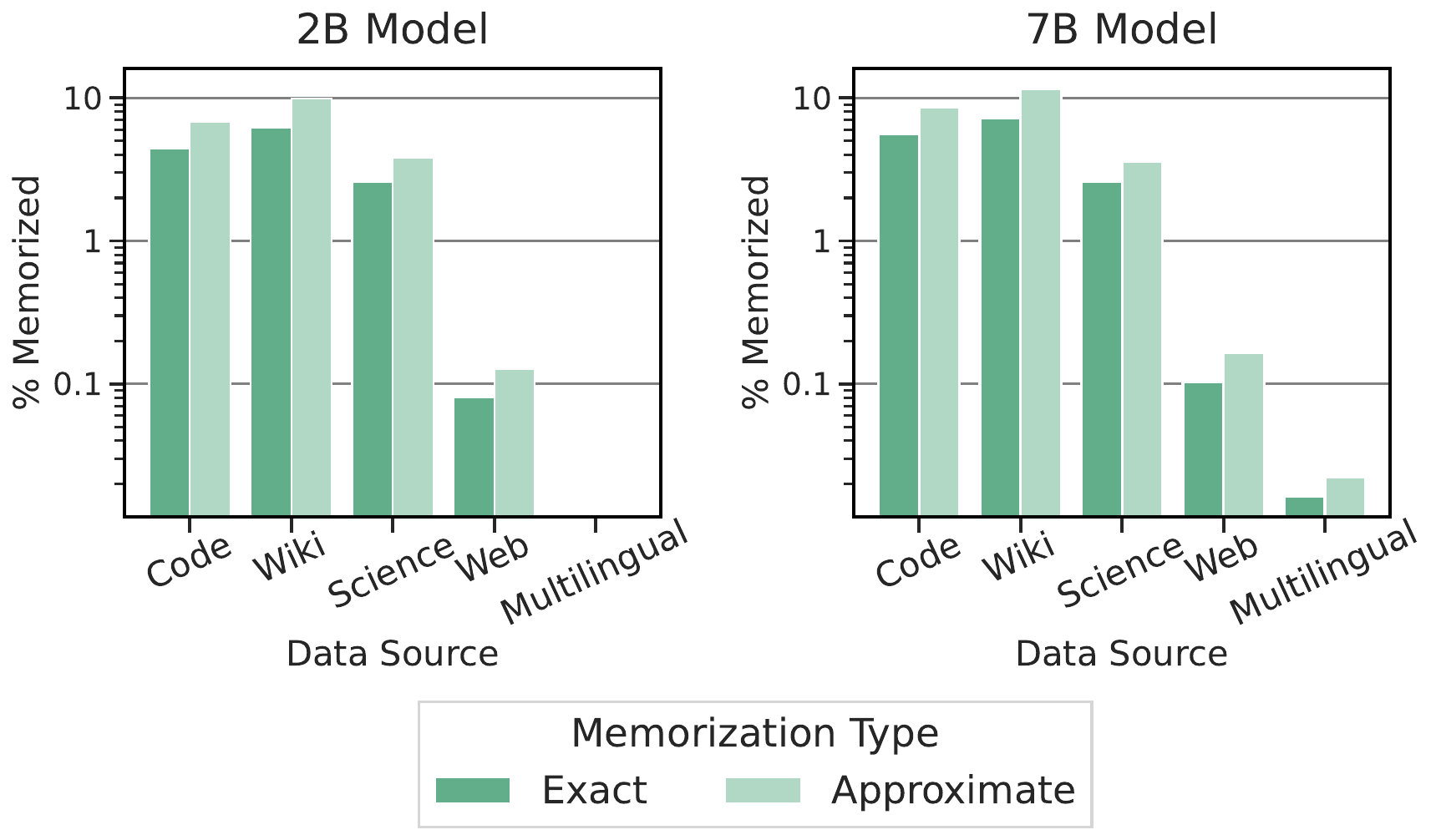}
    \caption{Comparing exact and approximate memorization.}
    \label{fig:mem-amem}
\end{figure}

\paragraph{Approximate Memorization} In Figure~\ref{fig:mem-amem}, we observe that roughly 50\% more data is approximately memorized (note the log scale) and that this is nearly consistent across each of the different subcategories over the dataset.

\section{Responsible Deployment}
\label{safety}

In line with previous releases of Google’s AI technologies \citep{geminiteam2023gemini,alphafoldRelease}, we follow a structured approach to responsible development and deployment of our models, in order to identify, measure, and manage foreseeable downstream societal impacts. As with our recent Gemini release, these are informed by prior academic literature on language model risks \citep{weidinger}, findings from similar prior exercises conducted across the industry \citep{anil2023palm}, ongoing engagement with experts internally and externally, and unstructured attempts to discover new model vulnerabilities.

\phantomsection
\subsection{Benefits}
We believe that openness in AI science and technology can bring significant benefits. Open-sourcing is a significant driver of science and innovation, and a responsible practice in most circumstances. But this needs to be balanced against the risk of providing actors with the tools to cause harm now or in the future. 

Google has long committed to providing broader access to successful research innovations (GraphCast, Transformer, BERT, T5, Word2Vec), and we believe that releasing Gemma into the AI development ecosystem will enable downstream developers to create a host of beneficial applications, in areas such as science, education and the arts. Our instruction-tuned offerings should encourage a range of developers to leverage Gemma’s chat and code capabilities to support their own beneficial applications, while allowing for custom fine-tuning to specialize the model’s capabilities for specific use cases. To ensure Gemma supports a wide range of developer needs, we are also releasing two model sizes to optimally support different environments, and have made these models available across a number of platforms (see \href{https://www.kaggle.com/models/google/gemma/frameworks/flax/variations/7b-it}{Kaggle} for details). Providing broad access to Gemma in this way should reduce the economic and technical barriers that newer ventures or independent developers face when incorporating these technologies into their workstreams.

As well as serving developers with our instruction-tuned models, we have also provided access to corresponding base pretrained models. By doing so, it is our intention to encourage further AI safety research and community innovation, providing a wider pool of models available to developers to build on various methods of transparency and interpretability research that the community has already benefited from \citep{evans,hendrycks}.

\subsection{Risks}
In addition to bringing benefits to the AI development ecosystem, we are aware that malicious uses of LLMs, such as the creation of deepfake imagery, AI-generated disinformation, and illegal and disturbing material can cause harm on both an individual and institutional levels \citep{weidinger}. Providing access to model weights, rather than releasing models behind an API, also raises new challenges for responsible deployment. 

First, we cannot prevent bad actors from fine tuning Gemma for malicious intent, despite their use being subject to Terms of Use that prohibit the use of Gemma models in ways that contravene our Gemma Prohibited Use Policy. However, we are cognizant that further work is required to build more robust mitigation strategies against intentional misuse of open models, which Google DeepMind will continue to explore both internally and in collaboration with the AI community.

The second challenge we face is protecting developers and downstream users against the unintended behaviours of open models, including generation of toxic language or perpetuation of discriminatory social harms, model hallucinations and leakage of personally identifiable information. When deploying models behind an API, these risks can be reduced via various filtering methods.

\vspace{-2mm}
\subsection{Mitigations}

Without this layer of defense for the Gemma family of models, we have endeavoured to safeguard against these risks by filtering and measuring biases in pre-training data in line with the Gemini approach, assessing safety through standardized AI safety benchmarks, internal red teaming to better understand the risks associated with external use of Gemma, and subjecting the models to rigorous ethics and safety evaluations, the results of which can be seen in \ref{tab:safety_auto_evals}.

While we've invested significantly in improving the model, we recognize its limitations. To ensure transparency for downstream users, we've published a detailed \href{https://ai.google.dev/gemma/docs/model_card}{model card} to provide researchers with a more comprehensive understanding of Gemma.

We have also released a Generative AI Responsible Toolkit to support developers to build AI responsibly. This encompasses a series of assets to help developers design and implement responsible AI best practices and keep their users safe.

The relative novelty of releasing open weights models means new uses, and misuses, of these models are still being discovered, which is why Google DeepMind is committed to the continuous research and development of robust mitigation strategies alongside future model development.

\vspace{-2mm}
\subsection{Assessment}
Ultimately, given the capabilities of larger systems accessible within the existing ecosystem, we believe the release of Gemma will have a negligible effect on the overall AI risk portfolio. In light of this, and given the utility of these models for research, auditing and downstream product development, we are confident that the benefit of Gemma to the AI community outweighs the risks described.  

\subsection{Going Forward}
As a guiding principle, Google DeepMind strives to adopt assessments and safety mitigations proportionate to the potential risks from our models. Although we are confident that Gemma models will provide a net benefit to the community, our emphasis on safety stems from the irreversible nature of this release. As the harms resulting from open models are not yet well defined, nor does an established evaluation framework for such models exist, we will continue to follow this precedent and take a measured and cautionary approach to open model development. As capabilities advance, we may explore extended testing, staggered releases or alternative access mechanisms to ensure responsible AI development.

As the ecosystem evolves, we urge the wider AI community to move beyond simplistic 'open vs. closed' debates, and avoid either exaggerating or minimising potential harms, as we believe a nuanced, collaborative approach to risks and benefits is essential. At Google DeepMind we're committed to developing high-quality evaluations and invite the community to join us in this effort for a deeper understanding of AI systems.

\section{Discussion and Conclusion}

We present Gemma, an openly available family of generative language models for text and code. Gemma advances the state of the art of openly available language model performance, safety, and responsible development.

In particular, we are confident that Gemma models will provide a net benefit to the community given our extensive safety evaluations and mitigations; however, we acknowledge that this release is irreversible and the harms resulting from open models are not yet well defined, so we continue to adopt assessments and safety mitigations proportionate to the potential risks of these models. In addition, our models outperform competitors on 6 standard safety benchmarks, and in human side-by-side evaluations.

Gemma models improve performance on a broad range of domains including dialogue, reasoning, mathematics, and code generation. Results on MMLU (64.3\%) and MBPP (44.4\%) demonstrate both the high performance of Gemma, as well as the continued headroom in openly available LLM performance.

Beyond state-of-the-art performance measures on benchmark tasks, we are excited to see what new use-cases arise from the community, and what new capabilities emerge as we advance the field together. We hope that researchers use Gemma to accelerate a broad array of research, and that developers create beneficial new applications, user experiences, and other functionality.

Gemma benefits from many learnings of the Gemini model program including code, data, architecture, instruction tuning, reinforcement learning from human feedback, and evaluations. As discussed in the Gemini technical report, we reiterate a non-exhaustive set of limitations to the use of LLMs. Even with great performance on benchmark tasks, further research is needed to create robust, safe models that reliably perform as intended. Example further research areas include factuality, alignment, complex reasoning, and robustness to adversarial input. As discussed by Gemini, we note the need for more challenging and robust benchmarks.

\clearpage

\section{Contributions and Acknowledgments}

\noindent\textbf{Core Contributors} \\
Thomas Mesnard \\
Cassidy Hardin \\
Robert Dadashi \\
Surya Bhupatiraju \\
Shreya Pathak \\
Laurent Sifre \\
Morgane Rivière \\
Mihir Sanjay Kale \\
Juliette Love \\
Pouya Tafti \\
Léonard Hussenot \\
Pier Giuseppe Sessa

\noindent\textbf{Contributors} \\
Aakanksha Chowdhery \\
Adam Roberts \\
Aditya Barua \\
Alex Botev \\
Alex Castro-Ros \\
Ambrose Slone \\
Amélie Héliou \\
Andrea Tacchetti \\
Anna Bulanova \\
Antonia Paterson \\
Beth Tsai \\
Bobak Shahriari \\
Charline Le Lan \\
Christopher A. Choquette-Choo \\
Clément Crepy \\
Daniel Cer \\
Daphne Ippolito \\
David Reid \\
Elena Buchatskaya \\
Eric Ni \\
Eric Noland \\
Geng Yan \\
George Tucker  \\
George-Christian Muraru \\
Grigory Rozhdestvenskiy \\
Henryk Michalewski \\
Ian Tenney \\
Ivan Grishchenko \\
Jacob Austin \\
James Keeling \\
Jane Labanowski \\
Jean-Baptiste Lespiau \\
Jeff Stanway \\
Jenny Brennan \\
Jeremy Chen \\
Johan Ferret \\
Justin Chiu \\
Justin Mao-Jones \\
Katherine Lee \\
Kathy Yu \\
Katie Millican \\
Lars Lowe Sjoesund \\
Lisa Lee \\
Lucas Dixon \\
Machel Reid \\
Maciej Mikuła \\
Mateo Wirth \\
Michael Sharman \\
Nikolai Chinaev \\
Nithum Thain \\
Olivier Bachem \\
Oscar Chang \\
Oscar Wahltinez \\
Paige Bailey \\
Paul Michel \\
Petko Yotov \\
Rahma Chaabouni \\
Ramona Comanescu  \\
Reena Jana \\
Rohan Anil \\
Ross McIlroy \\
Ruibo Liu \\
Ryan Mullins \\
Samuel L Smith \\
Sebastian Borgeaud \\
Sertan Girgin \\
Sholto Douglas \\
Shree Pandya \\
Siamak Shakeri \\
Soham De \\
Ted Klimenko \\
Tom Hennigan \\
Vlad Feinberg \\
Wojciech Stokowiec \\
Yu-hui Chen \\
Zafarali Ahmed \\
Zhitao Gong

\pagebreak
\noindent\textbf{Product Management} \\
Tris Warkentin \\
Ludovic Peran

\noindent\textbf{Program Management} \\
Minh Giang

\noindent\textbf{Executive Sponsors} \\
Clément Farabet \\
Oriol Vinyals \\
Jeff Dean \\
Koray Kavukcuoglu \\
Demis Hassabis \\
Zoubin Ghahramani \\
Douglas Eck \\
Joelle Barral \\
Fernando Pereira \\
Eli Collins

\noindent\textbf{Leads} \\
Armand Joulin \\
Noah Fiedel \\
Evan Senter

\noindent\textbf{Tech Leads} \\
Alek Andreev\begin{math}\dagger{}\end{math} \\
Kathleen Kenealy\begin{math}\dagger{}\end{math}
{\let\thefootnote\relax\footnote{\begin{math}\dagger{}\end{math} equal contribution.}}

\noindent\textbf{Acknowledgements} \\
Our work is made possible by the dedication and efforts of numerous teams at Google. We
would like to acknowledge the support from the following teams: Gemini, Gemini Safety, Gemini Infrastructure, Gemini Evaluation, Google Cloud, Google Research Responsible AI, Kaggle, and Keras.

\vspace{-3mm}
Special thanks and acknowledgment to Adrian Hutter, Andreas Terzis, Andrei Kulik, Angelos Filos, Anushan Fernando, Aurelien Boffy, Danila Sinopalnikov, Edouard Leurent, Gabriela Surita, Geoffrey Cideron, Jilin Chen, Karthik Raveendran, Kathy Meier-Hellstern, Kehang Han, Kevin Robinson, Kritika Muralidharan, Le Hou, Leonard Berrada, Lev Proleev, Luheng He, Marie Pellat, Mark Sherwood, Matt Hoffman, Matthias Grundmann, Nicola De Cao, Nikola Momchev, Nino Vieillard, Noah Constant, Peter Liu, Piotr Stanczyk, Qiao Zhang, Ruba Haroun, Seliem El-Sayed, Siddhartha Brahma, Tianhe (Kevin) Yu, Tom Le Paine, Yingjie Miao, Yuanzhong Xu, and Yuting Sun.

\bibliography{main}

\clearpage
\appendix
\onecolumn

\section{Gemma 1.0 IT results}
The core of the paper presents the results of the Gemma 1.1 IT models. We kept the results of the previous Gemma 1.0 IT models for comparison in this appendix. Side-by-side evaluations of Gemma 1.0 IT against Mistral 7b v0.2 can be found in table~\ref{tab:itv1}. Safety academic benchmark results of version 1.0 can be found in table~\ref{tab:safety_auto_evalsv1}.

\vspace{1cm}

\begin{table}[h]
    \setlength{\tabcolsep}{6pt}
    \centering
    \footnotesize
    \begin{tabular}{l r r}
    \toprule
    Model & Safety & Instruction Following \\
    \midrule
   \textbf{Gemma 7B IT} & \textbf{58\%} &  \textbf{51.7\%} \\
\tiny{\textit{95\% Conf. Interval}} & \tiny{[55.9\%, 60.1\%]} & \tiny{[49.6\%, 53.8\%]} \vspace{-0.05cm} \\
\tiny{\textit{Win / Tie / Loss}} & \tiny{42.9\% / 30.2\% / 26.9\%} & \tiny{42.5\% / 18.4\% / 39.1\%} \vspace{0.2cm} \\
   \textbf{Gemma 2B IT} & \textbf{56.5\%} &  41.6\% \\
\tiny{\textit{95\% Conf. Interval}} & \tiny{[54.4\%, 58.6\%]} & \tiny{[39.5\%, 43.7\%]}  \vspace{-0.05cm} \\
\tiny{\textit{Win / Tie / Loss}} & \tiny{44.8\% / 22.9\% / 32.3\%} & \tiny{32.7\% / 17.8\% / 49.5\%} \\
    \bottomrule
    \end{tabular}
    \caption{Win rate of Gemma 1.0 IT models versus Mistral 7B v0.2 Instruct with 95\% confidence intervals. We report breakdowns of wins, ties, and losses. Ties are broken evenly in the final win rate.}
    \label{tab:itv1}
\end{table}

\vspace{1cm}

\begin{table}[h!]
    \centering
    \begin{tabular}{l c c c  c c  c }
    \toprule
         & & Mistral v0.2 & \multicolumn{2}{c}{Gemma IT}\\
        \cmidrule(l{3pt}r{3pt}){3-3}\cmidrule(l{3pt}r{3pt}){4-5}
Benchmark & metric  & 7B* & 2B & 7B \\
      \midrule
RealToxicity & avg & 8.44 & \textbf{6.86} & 7.90 \\
BOLD &  & 46.0 & 45.57 & \textbf{49.08}  \\
CrowS-Pairs & top-1 & 32.76 & 45.82 & \textbf{51.33}  \\
BBQ Ambig & 1-shot, top-1 & \textbf{97.53} & 62.58 & 92.54 \\
BBQ Disambig & top-1 & \textbf{84.45} & 54.62 & 71.99 \\
Winogender & top-1 & \textbf{64.3} & 51.25 & 54.17 \\
TruthfulQA & & \textbf{48.54} & 31.81 & 44.84\\
Winobias 1\_2 &  & \textbf{65.72} & 56.12 & 59.09 \\
Winobias 2\_2 &  & 84.53 & 91.1 & \textbf{92.23} \\
Toxigen &  & 61.77 & \textbf{29.77} & 39.59 \\
\bottomrule
\end{tabular}
    \caption{Safety academic benchmark results of Gemma 1.0 IT models, compared to similar size open models. Evaluations run by us. Note that due to restrictive licensing, we were unable to run evals on LLaMA-2; we do not report previously-published numbers for LLaMA-2 on TruthfulQA, because we use different, non-comparable evaluation set-ups: we use MC2, where LLaMA-2 uses GPT-Judge.
    }
    \label{tab:safety_auto_evalsv1}
\end{table}

\end{document}